\documentclass[letterpaper, 10 pt, conference]{ieeeconf}  

\IEEEoverridecommandlockouts                              
\IEEEoverridecommandlockouts

\usepackage{cite}
\usepackage{amsthm}  
\usepackage{amsmath,amssymb,amsfonts}
\usepackage{graphicx}
\usepackage{textcomp}
\usepackage{xcolor}
\def\BibTeX{{\rm B\kern-.05em{\sc i\kern-.025em b}\kern-.08em
    T\kern-.1667em\lower.7ex\hbox{E}\kern-.125emX}}

\usepackage[utf8]{inputenc} 
\usepackage[T1]{fontenc}    
\usepackage[hidelinks]{hyperref}       
\usepackage{url}            
\usepackage{booktabs}       
\usepackage{nicefrac}       
\usepackage{wrapfig,lipsum,booktabs}

\usepackage[american]{babel}

\usepackage{todonotes}
\usepackage{xspace}
\usepackage[linesnumbered,ruled,noend]{algorithm2e}

\usepackage{mathtools}

\usepackage{amsmath}
\usepackage{cite}
\usepackage{amsmath,amssymb,amsfonts}
\usepackage{bbm}
\usepackage{subcaption}

\newenvironment{proof}{\noindent \textit{Proof.}}{\hfill$\square$}

\newenvironment{proofsketch}{\noindent \textit{Proof Sketch.}}{\hfill$\square$}

\makeatletter
\newcommand{\removelatexerror}{\let\@latex@error\@gobble}
\makeatother

\usepackage{algpseudocode}

\newcommand{\pX}{\mathbf{x}}
\newcommand{\pv}{\mathbf{v}}
\newcommand{\pW}{\mathbf{w}}
\newcommand{\pU}{\mathbf{u}}
\newcommand{\pz}{\mathbf{z}}

\newcommand{\dyn}{f(\pX_{k}, \pU_{k}) }

\newcommand{\up}[1]{\overline{#1}}
\newcommand{\low}[1]{\underline{#1}}

\newcommand{\unsafe}{\mathrm{u}}
\newcommand{\safe}{\mathrm{s}}

\newcommand{\Xs}{X_{\safe}}

\newcommand{\mean}{\hat{f}}

\newcommand{\kernel}{\kappa}
\newcommand{\std}{\sigma}

\newcommand{\GInput}{\mathrm{X}}
\newcommand{\Goutput}{\mathrm{Y}}
\newcommand{\dataset}{D}

\newcommand{\gpcov}{K}

\newcommand{\B}{\mathcal{B}}

\newcommand{\pw}{\mathbf{w}}
\newcommand{\px}{\mathbf{x}}

\newcommand{\reals}{\mathbb{R}}
\newcommand{\naturals}{\mathbb{N}}
\newcommand{\naturalszero}{\mathbb{N}_{\geq 0}}

\renewcommand{\P}{\mathcal{P}}

\newtheorem{problem}{Problem}
\newtheorem{proposition}{Proposition}
\newtheorem{theorem}{Theorem}
\newtheorem{assumption}{Assumption}
\newtheorem{definition}{Definition}
\newtheorem{remark}{Remark}
\newtheorem{corollary}{Corollary}

\title{\LARGE \bf 
Data-Driven Permissible Safe Control with Barrier Certificates
}
\author{
Rayan Mazouz$^{*1}$, John Skovbekk$^{*1}$, Frederik Baymler Mathiesen$^2$, 
Eric Frew$^1$, \\
Luca Laurenti$^2$, and Morteza Lahijanian$^1$%
\thanks{$^*$Equal contributions.}
\thanks{$^1$Authors are with the University of Colorado Boulder.}
\thanks{$^2$Authors are with Delft University of Technology.}
}

\begin{document}

\maketitle
\thispagestyle{plain}
\pagestyle{plain}

\maketitle

\begin{abstract}
This paper introduces a method of identifying a maximal set of safe strategies from data for stochastic systems with unknown dynamics using  barrier certificates. 
The first step is learning the dynamics of the system via Gaussian process (GP) regression and obtaining probabilistic errors for this estimate.
Then, we develop an algorithm for constructing piecewise stochastic barrier functions to find a maximal permissible strategy set using the learned GP model, which is based on sequentially pruning the worst controls until a maximal set is identified. 
The permissible strategies are guaranteed to maintain probabilistic safety for the true system.
This is especially important for learning-enabled systems, because a rich strategy space enables additional data collection and complex behaviors while remaining safe.
Case studies on linear and nonlinear systems demonstrate that increasing the size of the dataset for learning the system grows the permissible strategy set.
\end{abstract}

\section{Introduction}
\label{sec:intro}

In an era defined by the increasing integration of Artificial Intelligence (AI) into systems, ensuring the safety of stochastic systems with black-box components has become a major concern. These systems, characterized by uncertain and unpredictable dynamics, are ubiquitous across various domains, from autonomous vehicles \cite{shalev2017formal} to surgical robotics \cite{guiochet2017safety}. 
For such systems, the notion of the \emph{safety invariant set}, which represents regions of the system's state space where non-myopic safety constraints are guaranteed to be satisfied, 
emerges as a fundamental concept in providing safety guarantees. 
Particularly, identifying the set of \emph{permissible control strategies}, within these invariant sets holds critical importance, as it provides a comprehensive understanding of the system's operational boundaries and enables complex behaviors while guaranteeing safety. 
However, computation of such a strategy set poses a major challenge, especially if the system is unknown due to, e.g., AI or black-box components. In this work, we aim to provide a data-driven method for identifying a maximal set of permissible strategies that guarantee an unknown system remains inside a safe set.

Our approach is based on stochastic barrier functions (SBFs) \cite{SANTOYO2021109439, yu2023safe}. These functions provide a systematic method to bounding the system's behavior within a safe set, even in the presence of uncertainties or disturbances.  
We also utilize Gaussian process (GP) regression to learn the unknown dynamics, which enable probabilistic bounding of the learning error~\cite{berkenkamp2017safe,chowdhury2017kernelized,skovbekk2023formal}. 
Our key insight in dealing with the computational challenge of identifying permissible strategies is that local treatment of the uncertainty (namely, stochasticity and learning error) is needed, and a particular form of functions that enables such a treatment is piecewice (PW).  Hence, we employ the recent results in PW-SBFs~\cite{mazouz2024piecewise} to reason about subsets of controls.  Particularly, we show a formulation of PW-SBFs that allow admissibility assessment of local control subsets for each compact set of states via linear programming, achieving computational efficiency.
Based on these results, we propose an algorithm that, given a set of input-output data on the dynamics, safe and initial sets, and a lower-bound on the safety probability, returns a maximal permissible strategies set by iteratively removing inadmissible control sets until a fixed-point is reached.  We show soundness of the algorithm.  Our evaluations validate the theoretical guarantees.  They also illustrate that the approach is general and works on both unknown systems with linear and nonlinear stochastic dynamics.

In short, this work makes the following contributions:
\begin{itemize}
    \item a data-driven framework for computation of a maximal set of permissible strategies,
    \item extension of SBFs to provide safety control invariant sets for GP regressed models on continuous control sets,
    \item and a series of case studies on both linear and nonlinear systems that demonstrate the efficacy of the method and validate the theoretical guarantees.
\end{itemize}

\subsection{Related Work}

Probabilistic safety invariance is an essential property to enforce the safety of stochastic systems~\cite{wang2019data,gao2020computing,griffioen2023probabilistic,yu2023safe}.
For Markovian systems, these invariant sets can be found using linear programs (LPs) and mixed integer LPs for finite and infinite horizon probabilistic invariance, respectively~\cite{gao2020computing}. 
The sound application of these program requires knowledge of the transition kernel, which is not available for unknown systems as in our setting.
While a sampling-based procedure for probabilistic invariant sets of deterministic, stable black-box systems is available~\cite{wang2019data}, it does not admit controlled stochastic systems that may be inherently unstable. 

Gaussian process (GP) regression is notable for its flexibility in learning unknown systems and quantifying the uncertainty in safety-critical settings~\cite{berkenkamp2017safe,lederer2019local,jackson2021formal,griffioen2023probabilistic}. 
In addition to learning and estimating safe and stable regions for unknown systems~\cite{berkenkamp2017safe,lederer2019local}, they have been used to learn state-dependent modelling uncertainties for LTI control systems to subsequently find probabilistic invariant sets under a fixed feedback controller~\cite{griffioen2023probabilistic}. 
However, none of the previous can identify the full set of controls that leads to safety invariance over a prescribed horizon.

Stochastic Barriers Functions (SBFs) can provide bounds on the probability of remaining safe from an initial set~\cite{SANTOYO2021109439, yu2023safe}, even in the case a system is learned from data~\cite{jagtap2020formal,wajid2022formal}. 
Synthesis based on GP regression over states and controls can provide a single strategy with a lower-bound of safety~\cite{wajid2022formal}. 
In addition, SBFs can be used to find invariant sets for uncontrolled systems with bounded disturbances over an infinite horizon~\cite{yu2023safe}. Commonly, SBF synthesis relies on sum-of-squares optimization \cite{SANTOYO2021109439, mazouz2022safety} 
or neural barrier training \cite{mathiesen2022safety, dawson2023safe}. While both methods are deemed useful, they each have their own limitations. A novel and more efficient formulation relies on piecewise (PW) SBFs, overcoming some of these major limitations \cite{mazouz2024piecewise}. 
In this work, we adapt and extend PW-SBFs to enable computation of permissible controls for GP regressed models.

\label{sec:prelims}
\section{Problem Formulation}

Consider a stochastic dynamical system 
with dynamics 
\begin{equation}
\begin{split}
    \label{eq:system}
     &\pX_{k+1} = \dyn + \pw, \\
     k\in \naturals&, \quad \px_k \in \reals^n,  \quad \pU_{k} \in U\subset \reals^m, 
     \end{split}
\end{equation}
where $U$ is a compact set of controls, $f:\reals^n \times U \to \reals^n$ is the vector field,
and
$\pw$ represents noise taking values in $W\subseteq \reals^n$ and is assumed to be an independent and identically distributed (i.i.d.) sub-Gaussian random variable with known probability density function $p_{\pw}$. 
The choice of the control $\pU_k \in U$ is performed by a stationary (feedback) \textit{control strategy} $\pi:\mathbb{R}^n \to U$.  We denote the set of all control strategies by $\Pi$.

In this paper, we assume that vector field $f$
is \emph{unknown}.
Instead, we consider a given dataset $D= \{(x_i, u_i, x'_i)\}_{i=1}^M$, consisting of $M$ input-output i.i.d. samples from System \eqref{eq:system},  such that $x'_i=f(x_i,u_i) + w_i$, where $w_i$ is a realization of noise $\pw$.
We aim to learn $f$ using $D$.
To that end, we impose the following assumption, which ensures $f$ is a well-behaved analytical function on a compact set. 

\begin{assumption}[RKHS Continuity]
    \label{assume:gp}
     Let a compact set $Y \subset \mathbb{R}^n\times U$ and $\kappa : \mathbb{R}^{n+m} \times \mathbb{R}^{n+m} \to \mathbb{R}$ be a given kernel. Further, define $H_\kappa(Y)$ as the reproducing kernel Hilbert space (RKHS) of functions over $Y$ corresponding to $\kappa$ with norm $\| \cdot \|\kappa$ \cite{srinivas2012information}. Then, for each $i \in \{1, \ldots, n\}$, $f^{i} \in H_{\kappa}(Y)$ and for a constant $C_i > 0$, $\| f^{i} \|_\kappa \leq C_i$, where $f^{i}$ is the $i$-th component of $f$.
\end{assumption}

Assumption \ref{assume:gp} is standard in the literature and guarantees that $f$ can be learned via Gaussian processes (GPs) with an appropriate kernel. For instance, in the case where $\kernel$ is a universal kernel, such as the squared exponential kernel, it includes a class of functions that is dense w.r.t. the continuous functions over a compact set \cite{steinwart2001influence, gpbook}. 


Given these sources of uncertainty, we focus on probabilistic analysis of System~\eqref{eq:system}.  We begin by defining a probability measure over the trajectories of the system.
For a measurable set $X \subset \reals^n$, the one-step \textit{transition kernel} under a given strategy $\pi \in \Pi$ is 
\begin{equation}
\label{eq:transition_kernel}
    T(X\mid x,\pi):= \int_{\mathbb{R}^{\mathrm{w}}} \mathbf{1}_X (f(x,\pi(x)) + w)p_{\mathbf{w}}(w)dw,
\end{equation}
where $\mathbf{1}_X$ is the indicator function for $X$ such that
$\mathbf{1}_X(x) = 1$ if $x \in X$ and $\mathbf{1}_X(x) = 0$ if $x \not\in X$~\cite{bertsekas2004stochastic}.
Then, for a time horizon $N\in \mathbb{N}$, a strategy $\pi,$ and an initial condition $x_0 \in \mathbb{R}^n$, we consider the probability measure $\Pr$ over the trajectories of System~\eqref{eq:system} such that for  measurable sets $X_0, X_k \subseteq X$, it follows that 
\begin{equation}
    \label{eq:prob measure}
    \begin{aligned}
        &\Pr[\pX_0\in X_0] =  \mathbf{1}_{X_0}(x_0), \\
        &\Pr[\pX_k\in X_k   \mid \pX_{k-1}=x, \, \pi] =  T^{}(X_k \mid x,\pi).
    \end{aligned}
\end{equation}


\begin{definition}[Probabilistic Safety]
    Given a measurable safe set $\Xs \subset \reals^n$ and initial set $X_0 \subseteq \Xs$, the \emph{probabilistic safety} of System~\eqref{eq:system} under a control strategy $\pi$ for $N$ time steps is defined as
    \begin{multline*}
        \label{eq: safety probaility}
        P_s(X_s,X_0,N,\pi) = \\
        \inf_{x_0 \in X_0} \Pr[\px_k \in X_s \; \forall k \leq N \mid \px_0 = x_0, \pi].
    \end{multline*}
\end{definition}




It is enough to find a single strategy that satisfies a threshold on $P_s$ to claim that a system is safe.
However, identifying multiple strategies that guarantee a minimum $P_s$ can unlock more complex behaviors for the system.  
This is especially important for learning-enabled systems (e.g., the one considered in this work) because under such strategies more data can be collected in a safe manner to improve the learning model.
To this end, we define permissible strategies.

\begin{definition}[Permissible strategy]
    \label{def:admissble strategy}
    Given a safety threshold $p$, control strategy $\pi \in \Pi$ is called \emph{permissible} if 
    $$P_s(X_s,X_0,N,\pi) \geq p.$$
\end{definition}


Our goal is to find a maximal set of permissible strategies for System~\eqref{eq:system}.





\begin{problem}
\label{prob:1}
    Given dataset $D$ obtained from System~\eqref{eq:system}, safe set $X_s \subset \reals^n$, initial set $X_0 \subseteq X_s$, time horizon $N$, and safety probability threshold $p$, find a maximal set
    of permissible strategies $\Pi_s \subseteq \Pi$ such that every $\pi \in \Pi_s$ is permissible, i.e., 
    $$P_s(X_s,X_0,N,\pi) \geq p \qquad \forall \pi \in \Pi_s.$$
\end{problem}


Problem~\ref{prob:1} poses several challenges.  Firstly, it seeks a maximal set of permissible strategies for an unknown System~\eqref{eq:system}, which itself is a stochastic process.  Secondly, control set $U \subset \reals^m$ is continuous, making admissibility analysis of $\pi$ difficult due to its range being an uncountable, infinite set.
Furthermore, set $X_s$ can be non-convex, possibly leading to an overall non-convex optimization problem.

\begin{remark}
\label{remark:1}
    Note that a method that solves Problem 1 can also compute the probabilistic control invariant set.
    This set consists of states from which, if the system is initialized in it, it is possible to control the system to remain within the set under probabilistic guarantees. 
    The advantage of Problem~\ref{prob:1} is that it enables the control invariant set to be coupled with a maximal set of permissible strategies.
    
\end{remark}

\textbf{Approach overview. \quad}
Our approach to Problem~\ref{prob:1} is based on GP regression and stochastic barrier functions (SBFs).  
GP regression allows learning of the dynamics with formal learning-error quantification, and 
SFBs provide a formal methodology to prove invariance of stochastic systems. 
To address issues pertaining to 
continuity of $U$ and
non-convexity of $\Xs$, we utilize the recently-developed techniques for \textit{piecewise constant} (PWC) SBFs
\cite{mazouz2024piecewise}. 
We show that based on these functions, an inner-approximation of the set of permissible strategies can be obtained for System~\eqref{eq:system}.

\label{sec:prob}
\section{Preliminaries}

In this section, we provide a brief background on 
Gaussian process regression 
and PWC-SBFs,
which are core to our framework.

\subsection{Gaussian Process Regression}\label{sec:GPs}
A Gaussian process (GP) is a collection of random variables $\GInput$, any finite subset of which are jointly Gaussian~\cite{gpbook}.
Here, we describe GPs of single-dimension output but note that multidimensional outputs are similar.
GPs are often interpreted as distributions over a function space, which are defined completely by a mean function $\mean:\GInput\to\reals$ and covariance (kernel) function $\kernel:\GInput\times \GInput\to\reals$.
GP regression is a Bayesian approach to conditioning a prior GP model $(\mean_0, \kernel_0)$ on a dataset $\dataset=(\GInput,\Goutput)$ to find a posterior model $(\mean_\dataset, \kernel_\dataset)$ defined as
\begin{align*}
  \mean_\dataset(x) &= \mean_0(x) + \kernel_0(x, \GInput)\gpcov^{-1}(\Goutput - \mean_0(\GInput)), \\
  \kernel_\dataset(x, x') &= \kernel_0(x, x') - \kernel_0(x, \GInput)\gpcov^{-1}\kernel_0(\GInput, x'),
\end{align*}
where 
$\kernel_0(\GInput, x) = [\kernel_0(\GInput_i,x), \dots, \kernel_0(\GInput_M, x)]^T$ (similarly defined vectors of both arguments), $\gpcov = \kernel_0(\GInput, \GInput) + \std^2 I$, and $\std^2$ is the noise variance on observations $\Goutput$.
This posterior is a distribution of functions that best describes the data $\dataset$. 
When the input points to the kernel are identical, $\kernel^{-1/2}_D(x,x) = \sigma_D(x)$.



\subsection{Piecewise Constant Stochastic Barrier Functions}
Consider a stochastic process $\pz_{k+1} = F(\pz_k, \pv),$ where state $\pz_{k} \in \reals^{n_z}$, noise $\pv \in \reals^v$ is a random variable with distribution $p_\pv$, and vector field $F:\reals^{n_z} \times \reals^v \to \reals^{n_z}$ is almost everywhere continuous.
For a measurable compact set $Z \subset \reals^{n_z}$, let $T_z(Z \mid z)$ and Pr be the stochastic transition kernel and probability measure for process $\pz_k$ defined similarly to \eqref{eq:transition_kernel} and \eqref{eq:prob measure}, respectively.


Further, consider partition $Z_1,...,Z_{K_z}$ of $Z$ such that $F$ is continuous over each $Z_i$, and for scalars $b_i \in \reals_{\geq 0}$ with  $i = \{ 1, ..., K_z \}$, define PWC function $B: \reals^{n_z} \to \reals_{\geq 0}$ as
\begin{equation}
    \label{eq:pwc}
    B(z) = 
    \begin{cases}
        b_i & \text{if } z \in Z_i\\
        1 & \text{otherwise.}
    \end{cases}
\end{equation}

\noindent
The following theorem establishes the conditions under which $B$ is a \emph{PWC-SBF} for process $\pz$.

\begin{theorem}[{\cite[Theorem 2]{mazouz2024piecewise}}]
    \label{th:pwa}
    PWC function $B(z)$ in \eqref{eq:pwc} is a stochastic barrier function for process $\pz$ with safe sets $Z \subset \reals^{n_z}$ and initial set $Z_0 \subseteq Z$, if $\forall i \in \{1,\ldots,K_z\}$, there exist scalars $\beta,\eta \geq 0$ such that
    \begin{subequations} 
        \begin{align}
            & b_{i} \leq \eta  \qquad \qquad  \qquad \qquad \qquad \qquad \forall i : Z_i \cap Z_0,
            \label{eq:pwb_cond2}
            \\
            & \sum_{j = 1}^{K_z} b_j \cdot T_z(Z_j \mid z) + T_z(\reals^{n_z} \setminus Z \mid z) \leq \nonumber \\
            & \qquad \qquad \qquad \qquad \qquad b_{i} + \beta \qquad \forall i, \, \forall z \in Z_i.  \label{eq:total_law_exp}
        \end{align}
    \end{subequations}
    Then, for a $N\in \naturalszero$, it follows that
    \begin{align}
        \Pr [\forall k\leq N, \; \pz_k \in  Z \mid \pz_0 \in Z_0] \geq 1-( \eta + \beta N). \label{eq:probabilityBarrierFunctions}
    \end{align}
\end{theorem}


\noindent The advantage of SBFs is that, by satisfying static conditions, they allow for probabilistic reasoning about stochastic dynamical systems without having to unfold their trajectories. 

\section{Piecewise Stochastic Barriers for Invariance via GP Regression}


In this section, we introduce our main approach.  
First, we show that how GP regression can be used to bound transition kernel $T$.
Then, we propose a method to use the kernel bounds with Theorem~\ref{th:pwa} to identify permissible strategy sets with PWC-SBFs and provide soundness guarantees.

\subsection{Learning Transition Kernel Bounds}
\newcommand{\distBound}{\epsilon}
\paragraph{Learning $f$} To begin, we learn function $f$ in System~\eqref{eq:system} using GP regression on the state and control spaces using dataset $D$.
For brevity, let
$$z = (x,u)  \text{ \quad with \quad } z\in Z=\reals^n\times U.$$
Then, we estimate each component of $f$ with a GP map $\mean_D^i: Z \to \reals$ for $i = \{1, \ldots, n\}$.  
Assumption~\ref{assume:gp} allows for uniform probabilistic error bounds of $\mean_D^i$~\cite{chowdhury2017kernelized}. 
Namely, for scalar $\distBound^i\geq 0$ for $1 \leq i \leq n$ and compact set $\tilde{Z} \subset Z$, 
\begin{equation}\label{eq:prob}
    \Pr\left[|f^{i}(z)-\mean_D^i(z)| \leq \distBound^{i} \right] \geq 1-\delta \qquad \forall  z \in \tilde{Z},
\end{equation}
where 
$\delta$ satisfies $\distBound^{i}=\alpha(\delta) \sigma^{i}_{D}(z)$ 
as defined in \cite[Theorem 2]{chowdhury2017kernelized}.
The term $\alpha(\delta)$ scales the posterior according to the desired probability threshold and the RKHS properties of the function $f$, including its RKHS norm that can be bounded using its Lipschitz continuity~\cite{jackson2021formal}.
This probabilistic error bound is the foundation for computing bounds on the transition kernel $T$.


\paragraph{Bounding the Transition Kernel}

Directly computing the transition kernel $T$ for every state in a region in Theorem~\ref{th:pwa} 
is computationally intractable.  
Instead, we aim to compute bounds of $T$ using $\mean_D^i$ by employing
techniques from~\cite{skovbekk2023formal,jackson2021formal}.
Consider a partition $X_1,\ldots,X_K$ of safe set $\Xs$ in $K$ compact sets, and partition $U_1,\ldots,U_L$ of control space $U$ in $L$ compact sets, i.e.,
\begin{equation*}
   \bigcup_{i=1}^K X_i = \Xs \quad \text{and} \quad \bigcup_{l=1}^L U_l = U 
\end{equation*}
all of which are non-overlapping.
The product of these partitions is a partition of the state-control space, i.e., $$Z_{il} = X_i\times U_l.$$
Then, the transition kernel from a state-control partition $Z_{il}$ to a compact set $X_j \subset \reals^n$ of states can be bounded by finding the extrema of the  transition kernel, 
\begin{equation}
    \label{eq:Tbounds}
    \begin{aligned}
        & \low{p}^{l}_{ij} \leq \min_{x,u\in Z_{il}} T(X_j\mid x, u ), \\ 
        & \up{p}^{l}_{ij} \geq \max_{x,u\in Z_{il}} T(X_j\mid x, u).
    \end{aligned}
\end{equation}

\newcommand{\Post}{Post}
To compute bounds on these extrema, we partition the domains of the uncertainty, i.e., those of the additive noise $W$ and the learning error $\reals_{\geq 0}^n$ in \eqref{eq:prob} \cite{skovbekk2023formal}.
This method seeks all the uncertainty values to guarantee entering (avoiding) the target set to compute a non-trivial lower (upper) bound of the extrema.
Let $Q=W\times\reals_{\geq 0}^n$ be the product space of the noise and learning error domains, and
let $\mathcal{C} = \{C_1,\dots,C_p\}$ be non-overlapping partitions of 
$Q$. 
Also, let the learned image, $\Post(Z_{il},C)$, be all of the points in $Z_{il}$ propagated through $\mean_D$ with all uncertainties in $C \in \mathcal{C}$ accounted for, i.e.,
$$
\Post(Z_{il},C) = \{\mean_D(z) + c \mid z\in Z_{il}, c \in C\}.
$$
Finally, the extrema of $T$ are bounded using the following proposition.
\newcommand{\Indy}{\mathbf{1}}
\begin{proposition}[{\cite[Theorem 1]{skovbekk2023formal}}]\label{prop:transition}
    Let $\mathcal{C}$ be a partition of the noise space and learning error domain $Q = W \times 
    \reals^n_{\geq 0}$ of System~\eqref{eq:system}, and $Z_{il}$ be a compact subset of $Z$. 
    Then, the one-step transition kernel to a compact set $X \subset \reals^n$ from $Z_{il}$ is bounded by
\begin{align*}
    \min_{x,u\in Z_{il}} T(X\mid x,u) &\geq \sum_{C\in \mathcal{C}} \Indy(\Post(Z_{il}, C) \subseteq X) \Pr(C),\\
    \max_{x,u\in Z_{il}} T(X\mid x,u) &\leq \sum_{C\in \mathcal{C}} \Indy(\Post(Z_{il}, C) \cap X = \emptyset) \Pr(C),
\end{align*}
where $\Indy(\cdot)$ returns one if the argument is true and zero otherwise. 
\end{proposition}

Proposition~\ref{prop:transition} bounds the transition kernel for any choice of uncertainty partition $\mathcal{C}$.
\begin{remark}
    As the uncertainty structure of System~\eqref{eq:system} is relatively simple (it is additive for both the noise and learning error), an optimal partition $\mathcal{C}$ can be found that gives rise to the tightest bounds of $T$ in a straightforward manner.
    In fact, computation of the tightest-possible bounds requires only three partitions of the uncertainty space in each dimension.  See~\cite{skovbekk2023formal} for more details.    
\end{remark}



\subsection{Permissible Strategy Set Synthesis}

Using PWC-SBFs and the bounds on $T$, we extend Theorem \ref{th:pwa} to find the maximal permissible strategy set $\Pi_s$ for System~\eqref{eq:system}.
Let $X_\unsafe = \reals^n \setminus \Xs$, and 
consider the lower and upper bounds on the transition kernel $\low{p}^{l}_{ij} $ and $\up{p}^{l}_{ij}$ in~\eqref{eq:Tbounds}.
Note that $\low{p}^l_{ij}$ and $\up{p}^l_{ij}$ can be computed via Proposition~\ref{prop:transition}.
Then, the set of all feasible values for $T(X_j \mid x,u)$ for all $x \in X_i$ and $u \in U_l$ is 
\begin{align}
        \mathcal{P}^{l}_i = \Big\{ & p^{l}_i = (p^{l}_{i1},\ldots,p^{l}_{iK},  p^{l}_{i\unsafe})  \in [0,1]^{K+1} \quad s.t. \nonumber\\
        & \qquad \sum_{j=1}^K p^{l}_{ij}+p^{l}_{i\unsafe} = 1, \nonumber\\
        & \qquad \low{p}^{l}_{ij} \leq p^{l}_{ij} \leq \up{p}^{l}_{ij} \;\;\; \forall j \in \{1,\ldots,K,\unsafe \} \Big \}.
        \label{eq:feasible tran prob}
\end{align}

The following theorem sets up an optimization problem for synthesizing permissible strategy set $\Pi_s$ using PWC-SBF.

\begin{theorem}[PWC-SBF for Permissible Strategies]
    \label{th:pwc-xinv}
    Consider System~\eqref{eq:system} with safe set $\Xs \subset \reals^n$, initial set $X_0\subseteq \Xs$, dataset $D$, and safety probability bound $p$.
    Given $K$-partitions of $X_\safe$, let the set of PWC functions $\B_K$ be in the form~\eqref{eq:pwc}, with $B_i(x) =  b_i \in \reals_{\geq 0}$, for every $i \in \{1,\ldots,K\}$. 
    Further, given $L$-partitions of $U$, let $\P^l = \prod_{i=1}^K \P_i^l$, where $\P_i^l$ is the set of feasible transition kernel in \eqref{eq:feasible tran prob}, with the bounds computed by Proposition~\ref{prop:transition}, for every $l \in \{1,\ldots,L\}$.
    Finally, let $B^* \in \mathcal{B}_K$ 
    be
    a solution to the following optimization problem 
    \begin{align}
        \label{eq:outer_opt}
        & B^* = \arg\min_{B \in \mathcal{B}_K} \; \max_{ \{\P^l\}_{l=1}^L}
        \; \eta + N \beta
    \end{align}
    subject to
    \begin{subequations} 
        \begin{align}
            &b_i \leq \eta &&  \forall i : X_i \cap X_0 \neq \emptyset, \label{eq:bj_initial}\\
            &  \sum_{j = 1}^K b_j \cdot p^{l}_{ij} + p^{l}_{i\unsafe} \leq  b_i + \beta_{i}^l \label{eq:bj_martingale} && \forall i,~ \forall l,~ \forall p_i^l \in \P^l_i,\\
            &0 \leq \beta_{i}^l \leq \beta, && \forall i,~ \forall l.
        \end{align}
    \end{subequations}
    Then, 
    \begin{enumerate}
        \item $B^*$ constitutes a stochastic barrier certificate for System~\eqref{eq:system} that guarantees safety probability $P_s(X_s,X_0,N,\pi)  \geq 1 -  (\eta + \beta N)$ for every strategy $\pi \in \Pi$;
        \item if $\beta_i^l \leq (1 - \eta - p)/N $, then every choice of $u \in U_l$ is permissible for every $x \in X_i$ of System~\eqref{eq:system}.
    \end{enumerate}
\end{theorem}
\begin{proofsketch}
    The optimization is an exact solution to the problem in Theorem~\ref{th:pwa}.
    Hence, it is straightforward that 
     $B^*$ is a proper stochastic barrier certificate, and that
    for every strategy $\pi \in \Pi$, $P_s(X_s,X_0,N,\pi)  \geq 1 -  (\eta + \beta N)$, where $\beta = § \max \beta_{i}^{l}$.
    Further, it follows that if $\beta_i^l \leq (1 - \eta - p)/N $, state-control partition $Z_{il}$ is rendered safe $ \forall (x,u) \in X_i\times U_l$. Since the transition kernel bounds of the learned system in~\eqref{eq:feasible tran prob} are contained in the distribution of the true system~\cite[Theorem 1]{jackson2021formal}, the probabilistic safety guarantees hold for System \eqref{eq:system}.
\end{proofsketch}

The above theorem provides a method of identifying permissible strategies, which requires solving a minimax optimization problem.  This problem includes bilinear terms, i.e., $b_j \cdot p^l_{ij}$ in Condition~\eqref{eq:bj_martingale}. Nevertheless, by introducing dual variables, the minimax problem becomes a simple linear program (LP) as shown in~\cite{mazouz2024piecewise}, which guarantees efficiency. 
Furthermore, following the same approach developed in~\cite{mazouz2024piecewise}, the resulting LP can be efficiently solved by a Counter Example Guided Synthesis (CEGS) approach or a gradient descent method.
Our evaluations in Section~\ref{sec:evaluations} use the CEGS approach.

\begin{algorithm}[t]
    \caption{PWC-SBF based Control Invariant Sets }
    \label{alg:overview}

    \SetKwInOut{Input}{Input} 
    \SetKwInOut{Output}{Output}
    
    \Input{Initial set $X_0$, state-control space $\mathcal{Z}$,
    time horizon $N$, feasible transition kernel sets $ \tilde{\mathcal{P}} = \{\P^l\}_{l=1}^L$, and probability threshold $p$. 
    }
    \Output{Permissible strategy set $\Pi_s$.} 
    
    \BlankLine    
    $\eta, \beta \gets  \textsc{Barrier}(X_0, \mathcal{Z}, N,  \tilde{\mathcal{P}} $)  \Comment{Theorem~\ref{th:pwc-xinv} }
    \label{alg:first_barrier}\\


    
    \While{$1-(\eta+N\beta) < p $
    }{



    
    $Z_{il}',\mathcal{P}_{i}^{l}  \gets $ Identify region with $\max \beta_i^l$ 
    \label{alg:identifyZ} \\


    $ \mathcal{Z} \gets $ \textsc{} $ \mathcal{Z} \setminus Z'_{il} $ \hfill\Comment{Remove worst region} \
    \label{alg:updatepairs}\\

    $\tilde{\mathcal{P}} \gets $ \textsc{} $\tilde{\mathcal{P}}\setminus \mathcal{P}_{i}^{l} $ \hfill \hfill\Comment{Remove kernel bounds} \label{alg:updatebounds} \\

    $\eta, \beta \gets  \textsc{Barrier}(X_0, \mathcal{Z}, N, \tilde{\mathcal{P}}$)  \Comment{Theorem~\ref{th:pwc-xinv}}
    \label{alg:barrier_update} \\

    \If{ all control sets removed for $X_i$  \label{alg:check_if_exists}}
    {
    \Return False
    }
    $\Pi_s \gets \{ X_i \to U_l \mid \forall (X_i,U_l) \in \mathcal{Z}\}\label{alg:sets}$ 
}
    \Return  $\Pi_s$ \label{alg:return}
\end{algorithm} 

From the results of Theorem~\ref{th:pwc-xinv}, we propose an algorithm that computes a maximal set of permissible strategies~$\Pi_s \subseteq \Pi$ w.r.t. to partitions of the safe set and control space.
An overview of the algorithm is shown in Alg.~\ref{alg:overview}. 
For convenience, define set $\mathcal{Z}$ as 
\begin{align*}
     \mathcal{Z} = \{Z_{il} \mid Z_{il} = X_i \times U_l, \; \forall i, \; \forall l\}  
\end{align*}    
The algorithm takes as input the initial set $X_0$, set $\mathcal{Z}$, time horizon $N$, the feasible bounds on the transition kernel $T$, and the desired probability of safety $p$. 
On Line \ref{alg:first_barrier}, the algorithm computes an initial PWC-SBF using Theorem~\ref{th:pwc-xinv}. 
Next, the while loop is instantiated, with a termination condition that the safety probability criteria $p$ must be met. 

Using the PWC-SBF, the worst state-control pair ${Z}'_{il}$ corresponding to $\max \beta_{i}^{l}$, along with the probability bounds $ {P}_i^{l}$,  are identified on Line \ref{alg:identifyZ}. 
These are subsequently removed from their respective sets on Lines \ref{alg:updatepairs} and \ref{alg:updatebounds}. 
 Intuitively, this means that
the algorithm sequentially prunes control region $U_{l}$ that is deemed not permissible for a given state partition $X_i$, corresponding to $\max \beta_{i}^{l}$.
Then, a new PWC-SBF is synthesized on Line \eqref{alg:barrier_update} using Theorem \ref{th:pwc-xinv}.
A check procedure on Line \eqref{alg:check_if_exists} is incorporated to make sure that for all regions $X_i$, there exists at least one control region $U_l$ inside updated space $\mathcal{Z}$. If the latter is true, the maximal set of permissible strategies $\Pi_s$ is updated accordingly on Line \eqref{alg:sets}. Otherwise, the algorithm terminates as \texttt{False}.
A successful algorithm  termination (Line 10) implies that the probability threshold $p$ is met.


\begin{corollary}
    Alg.~\ref{alg:overview} terminates in finite time. If it terminates with the return statement on Line~\ref{alg:return}, then it returns $\Pi_s$, which is guaranteed to include only permissible strategies.
\end{corollary}
The proof is a direct implication of Theorem~\ref{th:pwc-xinv}.

\begin{remark}
    Inline with Remark~\ref{remark:1}, note that Alg. \ref{alg:overview} can also compute the probabilistic control invariant set. This is established by running the algorithm for every $X_i \in \Xs$, $i \in \{1,\ldots,K\}$  as the initial set. 
\end{remark}
\section{Case Studies}
\label{sec:evaluations}
We demonstrate the efficacy of Alg.~\ref{alg:overview} by finding permissible strategy sets for linear and nonlinear systems learned from data. 
For each system,
we perform GP regression to estimate the system and then find the permissible sets. 
Each GP uses zero mean and squared exponential kernel priors.
To determine the quality of our solution, we also compute the permissible strategy set of the known linear system on the same partitions.

In all experiments, the initial state region of interest is $X_0 = [0.4,0.5]^2$, the safe set is $\Xs = [0.0~1.0]^2$, and the noise $\pW$ is Gaussian zero-mean with variance $0.01^2$ and zero covariance in each dimension. 
Permissible sets are constructed with Alg.~\ref{alg:overview} using a convergence threshold of $p = 1 - 1\times10^{-4}$.
Then, this set is used to generate bounds on the probability of safety for $N = 100$ time-step trajectories initialized from $X_0$.

The safety probability bounds are validated by Monte Carlo simulations randomly initialized in $X_0$ and propagated by random sampling of the controls from the permissible set.
To further validate the guarantees of the approach, adversarial trajectories are generated by choosing control inputs that drive the system towards exiting the safe set from both the full control set and the permissible set.

Our implementation of Alg.~\ref{alg:overview} is written in Julia, and all the computations were performed on an Apple M2 Pro MacBook with 16GB of RAM.

\begin{figure*}
    \centering
    \begin{subfigure}{\linewidth}
        \centering \includegraphics[width=0.66\linewidth]{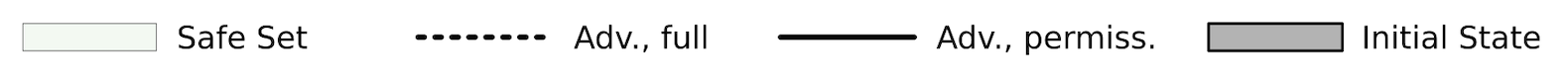}
    \end{subfigure}
    \begin{subfigure}{0.32\linewidth}
        \includegraphics[width=\linewidth]{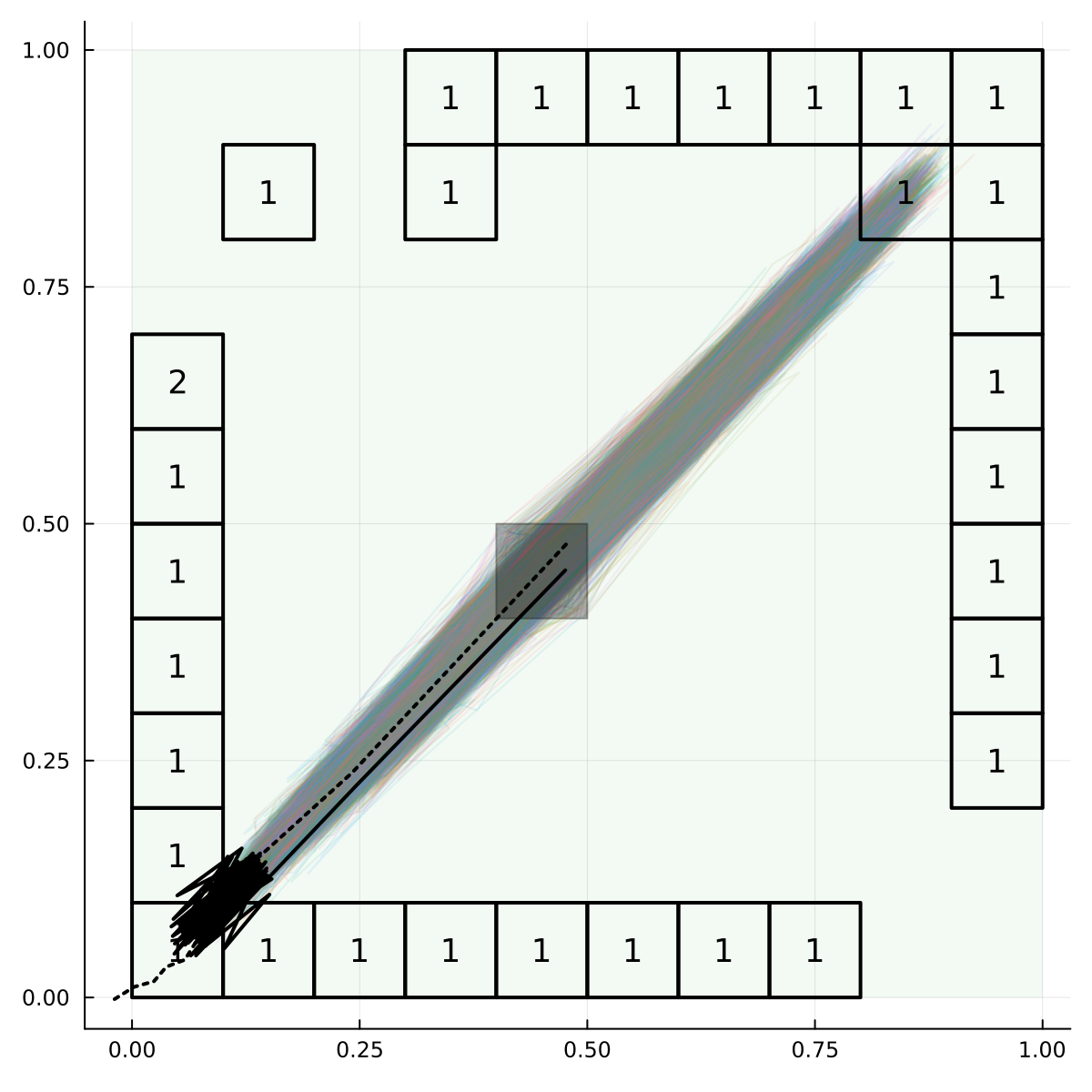}
        \caption{Known System - $P_\safe \geq 1- 5.83\times10^{-5}$}\label{fig:known1}
    \end{subfigure}
    \begin{subfigure}{0.32\linewidth}
        \includegraphics[width=\linewidth]{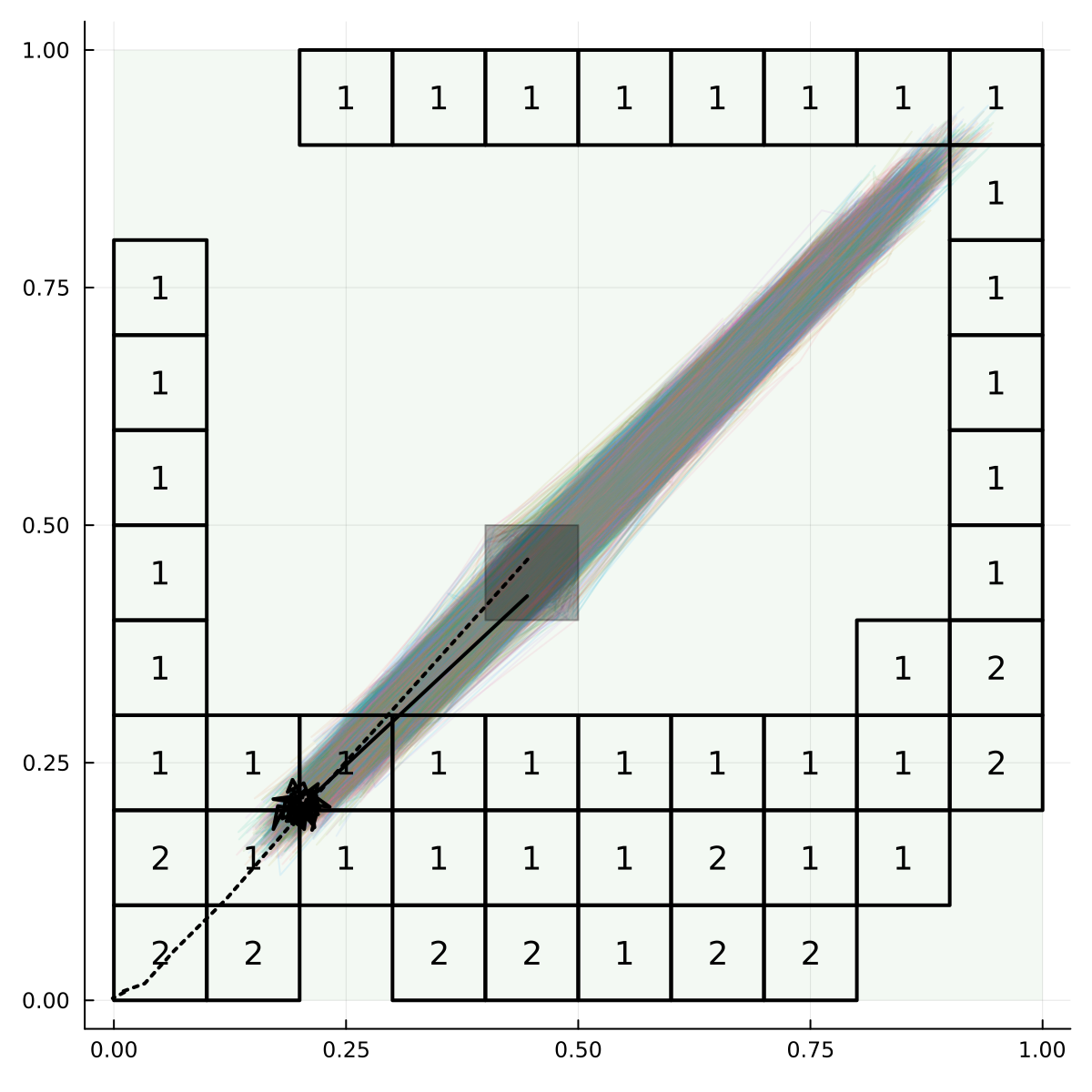}
        \caption{GP 500 - $P_\safe \geq 1- 2.39\times10^{-4}$}\label{fig:gp5001}
    \end{subfigure}
    \begin{subfigure}{0.32\linewidth}
        \includegraphics[width=\linewidth]{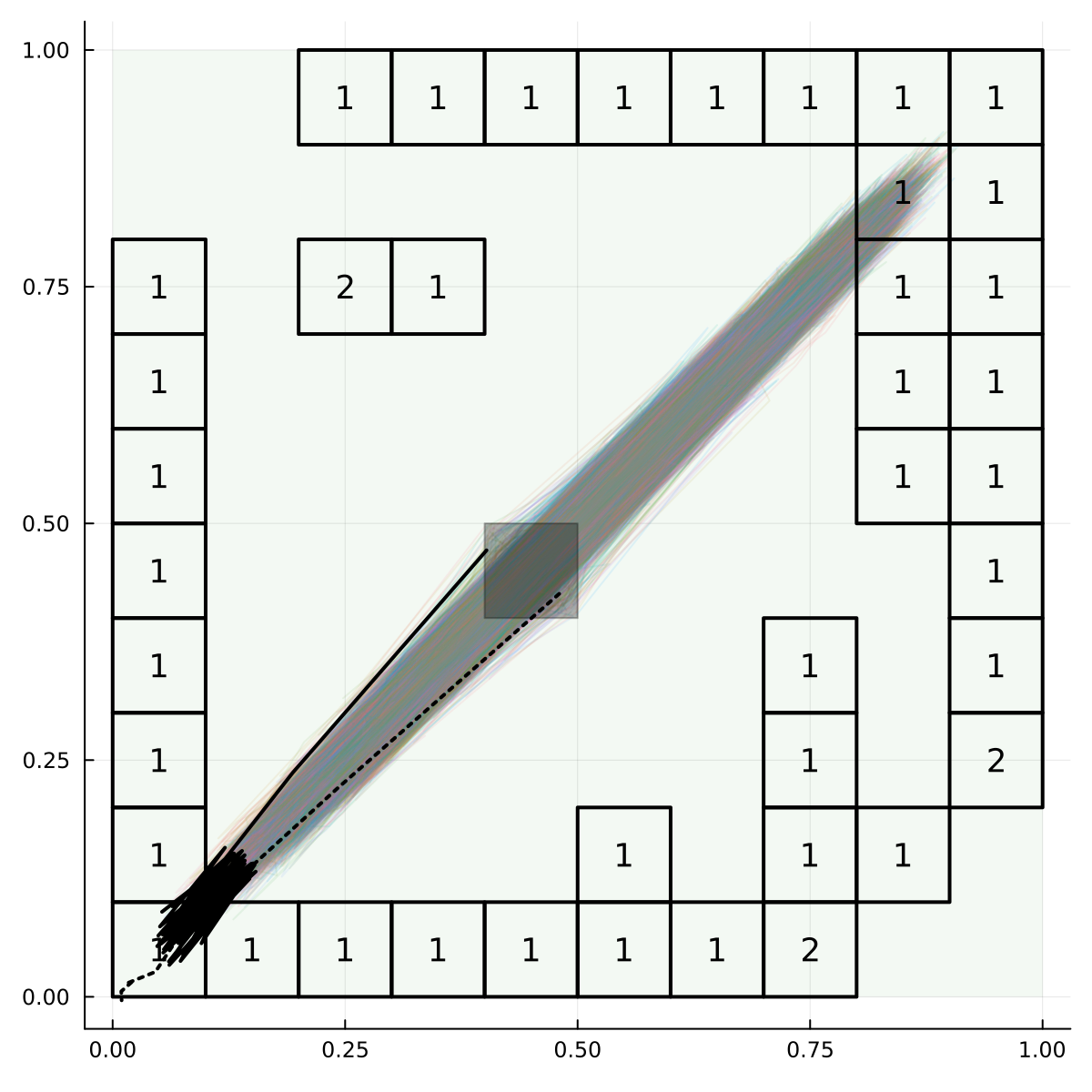}
        \caption{GP 2000 - $P_\safe \geq 1- 1.93\times10^{-5}$}\label{fig:gp20001}
    \end{subfigure}
    \caption{Results for the linear system case studies using the known system and two datasets. 
    Regions with \# of removed control regions and simulated trajectories over 100 steps are shown in each plot. Adversarial trajectories show permissible sets maintain safety.}
    \label{fig:results-1d}
\end{figure*}

\subsection{Linear System}
\label{sys:linear}
First, consider the 2D system $\pX_{k+1} = A\pX_k + B\pU_k + \pW_k$, where $A=0.5I$, $B = (1,1)^T$, and $u\in [0.0,0.5]$. 
From the initial set, this system primarily traverses around the diagonal $x_1=x_2$ line.
The safe set is discretized using a uniform grid with length 0.1, and the control range is partitioned into five intervals.
Computing each permissible set takes approximately 70-100 seconds for this system.

Figure~\ref{fig:results-1d} shows the safe and the initial sets.  
It also shows the final results for the known and learned systems.
The number inside each cell is the number of removed control sets.
In each case, we simulated 1000 trajectories using randomly sampled permissible strategies, which resulted in zero violations.  
These trajectories are shown in gray color.
Observe that, while the system can reach towards the edges of the safe set, the permissible strategy set ensures the system remains inside the safe set.

To further demonstrate the efficacy of the permissible strategy set, we compare trajectories that choose the most adversarial actions according to the full control set (shown in dashed black) and the permissible strategy set (shown in solid black).
Note that without restricting the strategies, the system can always be driven outside of the safe set.

Figures~\ref{fig:gp5001} and \ref{fig:gp20001} show the result using a GP regression with 500 and 2000 datapoints sampled from the system respectively. 
For the smaller dataset, more control partitions are removed from the permissible set, resulting in trajectories that cover a smaller range of the safe set. 
The learned model with 2000 data points results in an permissible set and samples that more resembles the known system, as fewer control partitions are removed.
The probabilities of exiting the safe set in 100 time steps is very small, and the learned system with 2000 datapoints has a tighter upper bound than the known system shown in Figure~\ref{fig:known1}.
This is likely due to convergence-related hyperparameters in the optimization procedure used to synthesize the PWC-SBF in Alg.~\ref{alg:overview}.

The richness of the permissible strategy set is indicated by the proportion of the control space it contains.
The permissible strategy set for a known system encompasses 93.6\% of the system's total possible actions, serving as a benchmark for learning-based methods. 
With 500 data points, the procedure recognizes 88.8\% of the entire action space, while incorporating up to 2000 data points enhances this figure to 91.2\%. 
This indicates that as more data becomes available to estimate the system, the permissible strategy set progressively approximates that of the known system.


\begin{figure*}
    \centering
    \begin{subfigure}{\linewidth}
        \centering \includegraphics[width=0.66\linewidth]{img/legend.png}
    \end{subfigure}
    \begin{subfigure}{0.32\linewidth}
        \includegraphics[width=\linewidth]{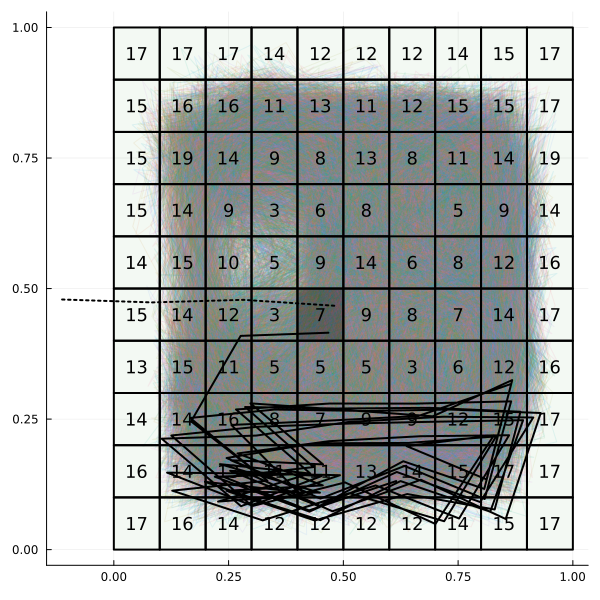}
        \caption{GP 500 - $P_\safe \geq 1- 8.96\times10^{-3}$}\label{fig:gp5002}
    \end{subfigure}
    \begin{subfigure}{0.32\linewidth}
        \includegraphics[width=\linewidth]{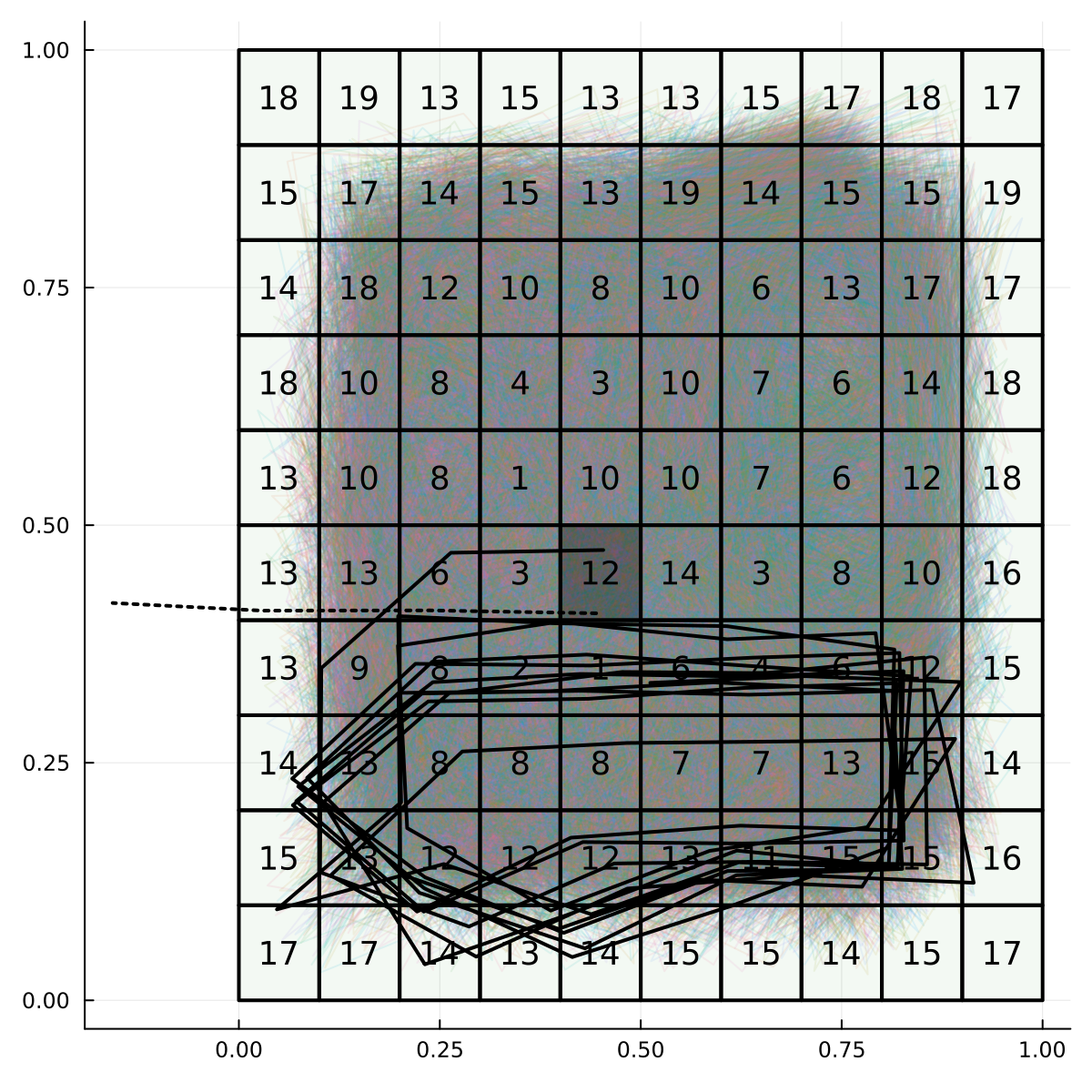}
        \caption{GP 1000 - $P_\safe \geq 1- 9.10\times10^{-3}$}\label{fig:gp10002}
    \end{subfigure}
    \begin{subfigure}{0.32\linewidth}
        \includegraphics[width=\linewidth]{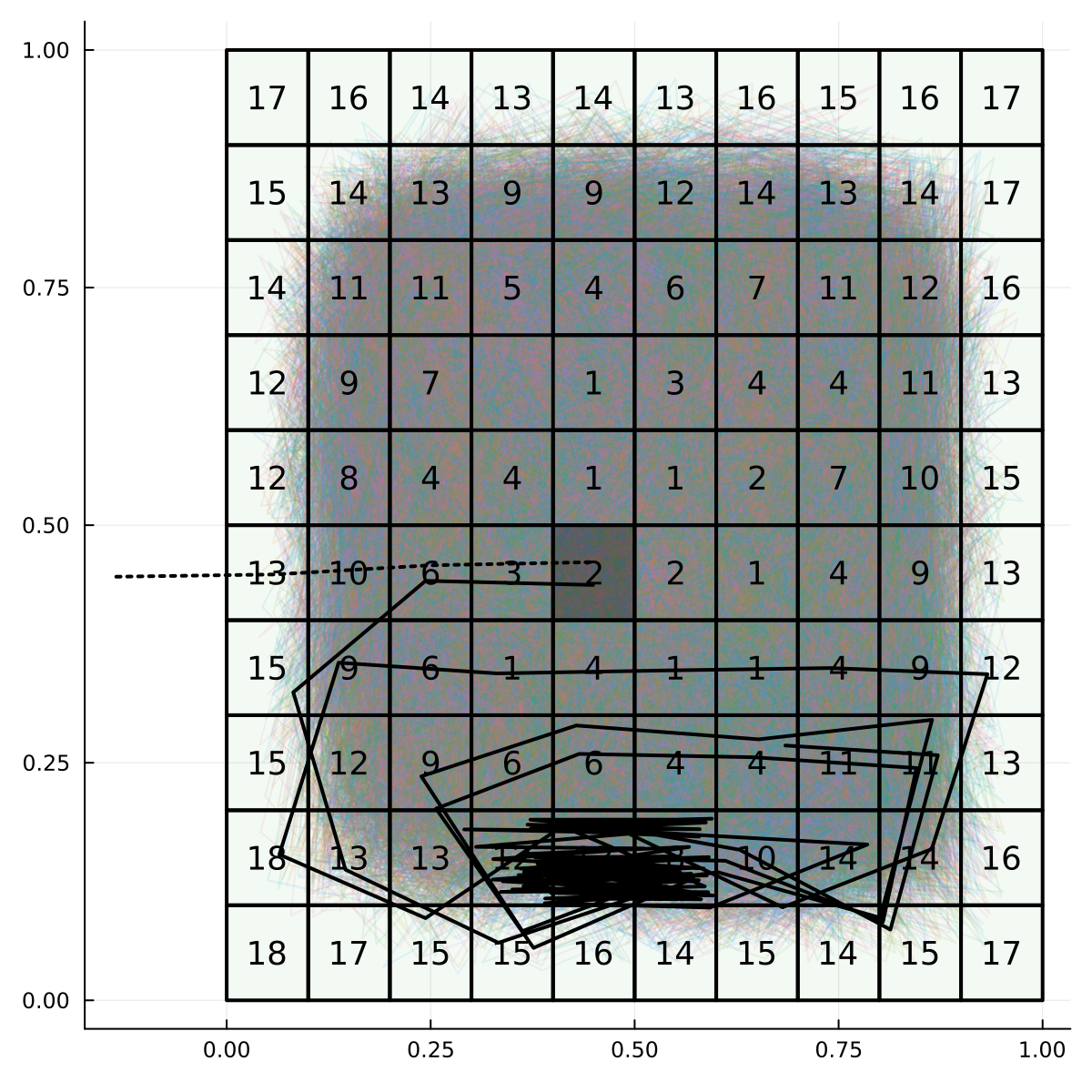}
        \caption{GP 1500 - $P_\safe \geq 1- 9.95\times10^{-3}$}\label{fig:gp15002}
    \end{subfigure}
    \caption{Results for the nonlinear system case studies using various sizes of datasets. Regions with \# of removed control regions and simulated trajectories over 100 steps are shown in each plot. Adversarial trajectories show permissible control sets maintain safety.}
    \label{fig:results-nl}
\end{figure*}

\subsection{Nonlinear System}

Consider the nonlinear dynamical system given by $\pX_{k+1} = \pX_k + 0.2[\cos(\pU_k),~\sin(\pU_k)]^T + \pW_k$, where $u\in[-\pi, \pi]$. 
This is a simplified version of a Dubin's car model, where the heading is set directly, allowing movement in any direction.
The state-space partition is the same as for the system in Section~\ref{sys:linear}, and the control range is partitioned into 20 intervals. 
The permissible set computation takes approximately 1 to 1.5 hours for each dataset.


Figure~\ref{fig:results-nl} illustrates the comparison of permissible strategy sets obtained for the system using datasets of 500, 1000, and 1500 data points.
In constructing the permissible sets, numerous control intervals that would lead the system to approach the safe set boundary are excluded. 
Despite the variations in dataset size, each permissible set constrains the probability of exiting the safe set within a similar bound. 

The resulting permissible sets for the system learned with 500 and 1000 datapoints maintained 39.5\% and 40.1\% of all available controls, respectively. 
With an additional 500 datapoints, the proportion of controls available rises to 49.5\%. 

Out of the 1000 sampled trajectories using random permissible strategies, none exited the safe set, and only the adversarial trajectory using the full control set violated safety.
The strict bound on $\beta$ during the synthesis of the permissible set removed a majority of the controls to ensure safety.
If the bound on $\beta$ were relaxed, then additional strategies could be added to the permissible set. 

\section{Conclusion}
In this work, we introduce a data-driven framework for the computation of a maximal set of permissible strategies for ensuring the safety of unknown stochastic systems. 
The unknown model with a continuous control space is regressed from data using Gaussian processes. 
Then, the regressed model is used to synthesize a piecewise stochastic barrier function, which in turn identifies 
theoretically-sound permissible control sets.
The safety guarantees are validated in case studies of linear and nonlinear systems. 
Future work includes the incorporation of other regression approaches, extensions to general temporal logic specifications, and applications to online safe exploration and learning.

\bibliographystyle{IEEEtran}
\bibliography{cite}
\end{document}